\documentclass[conference]{IEEEtran}
\IEEEoverridecommandlockouts
\usepackage{cite}
\usepackage{amsmath,amssymb,amsfonts}
\usepackage{algorithmic}
\usepackage{graphicx}
\usepackage{textcomp}
\usepackage{xcolor}
\usepackage{placeins}
\usepackage{hyperref}
\usepackage{multirow}
\usepackage{float}   
\usepackage{booktabs}
\def\BibTeX{{\rm B\kern-.05em{\sc i\kern-.025em b}\kern-.08em
    T\kern-.1667em\lower.7ex\hbox{E}\kern-.125emX}}

\usepackage{etoolbox}

\makeatletter
\pretocmd{\@maketitle}{%
    \noindent
    \parbox{\textwidth}{%
        \centering
        \footnotesize
        Accepted to the Workshop on Data Quality Aware, High-Performance,
        and Trustworthy AI Systems for Healthcare at IEEE/ACM CHASE 2026
    }
    \par\vspace{1.2em}
}{}{}
\makeatother
    
\begin{document}

\title{X-Beat: An Explainable Framework for ECG Image Classification\\\
}

%\title{X-Beat: A Benchmark and Interpretation Study of Transfer Learning for ECG Image Classification}

\author{
Mohammad Sadman Tahsin$^{1}$, 
Haitham Y. Adarbah$^{2 *}$\thanks{* Corresponding author: Haitham.Adarbah@tamuk.edu}, 
and Afzel Noore$^{3}$\\
\textit{AI-Based Autonomous System Research Lab}\\
\textit{Department of Electrical Engineering and Computer Science}\\
\textit{Texas A\&M University-Kingsville (TAMUK)}\\
Kingsville, TX, USA\\
$^{1}$Mohammad.Tahsin@students.tamuk.edu,
$^{2}$Haitham.Adarbah@tamuk.edu,
$^{3}$Afzel.Noore@tamuk.edu
}

%\and
%\and
%\and
%\and
%\and

\maketitle

\begin{abstract}
Accurate automated interpretation of electrocardiograms (ECGs) is essential for early detection of cardiac conditions such as myocardial infarction and rhythm abnormalities. However, many high-performing deep learning models remain difficult to deploy in clinical settings due to limited transparency and lack of reliability validation. In this work, we present X-Beat, an explainable and reliability-aware benchmark framework for ECG image classification designed to support trustworthy AI systems in healthcare. The proposed framework combines transfer learning with post-hoc explainability and systematic reliability evaluation across four cardiac classes: Abnormal Heartbeat, History of Myocardial Infarction, Myocardial Infarction, and Normal Heartbeat. Multiple ImageNet-pretrained CNN backbones, including EfficientNet-B0, ResNet-50, DenseNet-121, and MobileNetV3-Large, are evaluated under a unified training protocol. Beyond standard performance metrics, we incorporate Grad-CAM-based visual explanations together with additional analyses, including explanation stability, regional sensitivity, and confidence-based reliability assessment, to examine whether model predictions are supported by clinically meaningful evidence. Experimental results show that ResNet-50 achieves the best performance, reaching 91.94\% accuracy and a macro F1-score of 0.9098, with strong class separability (AUC up to 0.995). Explanation analyses indicate that the model primarily focuses on waveform-relevant regions, while reliability evaluation suggests that most incorrect predictions occur with lower confidence. Overall, this work provides a structured and reproducible benchmark for evaluating both predictive performance and explanation reliability in ECG image classification, contributing toward the development of trustworthy and interpretable AI components for clinical decision support systems.
\end{abstract}


\section{Introduction}

Electrocardiograms (ECGs) are widely used for detecting cardiac abnormalities such as myocardial infarction and rhythm disorders. In recent years, deep learning has shown strong potential for automated ECG interpretation, with prior studies reporting high performance for arrhythmia detection and related cardiac classification tasks \cite{sangha2022automated, BALOGLU201923}. However, many high-performing models remain difficult to trust because their decision process is not transparent, which limits their acceptance in medical sectors \cite{e23010018}. This issue is especially important in ECG analysis, where clinicians need evidence that model predictions are driven by meaningful waveform characteristics rather than irrelevant visual or dataset-specific patterns. This limitation is critical for the deployment of AI in connected healthcare systems, where models must be reliable, auditable, and aligned with clinical reasoning to support real-world clinical decision-making. 

Most prior ECG DL studies have focused on raw signal inputs, while image-based ECG classification remains comparatively less explored despite its practical relevance for scanned or printed ECG records \cite{sangha2022automated}. At the same time, explainability methods such as Gradient-weighted Class Activation Mapping (Grad-CAM) have become increasingly important for examining whether convolutional neural networks attend to clinically relevant regions during prediction \cite{goettling2024xecgarch}. These developments motivate a compact and explainability-aware evaluation of transfer learning for ECG image classification under limited-data conditions.

In this work, we study explainable ECG image classification using a publicly available dataset containing Normal Heartbeat, Abnormal Heartbeat, Myocardial Infarction, and History of Myocardial Infarction samples. The proposed pipeline includes image preprocessing, transfer learning with multiple pretrained CNN backbones, and post-hoc visual explanation using Grad-CAM. Four ImageNet-pretrained models, EfficientNet-B0, ResNet-50, DenseNet-121, and MobileNetV3-Large, are evaluated under a unified fine-tuning protocol, and their performance is compared using accuracy, macro-F1, ROC analysis, and confusion-matrix-based error analysis. The main contributions of this paper are as follows:

\begin{itemize}
    \item We provide a unified benchmark of pretrained CNN backbones for four-class ECG image classification on a limited image dataset.

    \item We combine standard performance reporting with class-wise and error-focused analysis to better characterize model behavior.

    \item We use Grad-CAM together with supplementary empirical checks to examine whether the learned predictions are supported by visually meaningful ECG regions.
\end{itemize}

%The experimental results show that ResNet-50 achieves the best overall performance, reaching 91.94\% accuracy and 0.9098 macro-F1 on the test set. The results further show strong separability for Myocardial Infarction and Normal Heartbeat, while the main difficulty lies in distinguishing History of Myocardial Infarction from Normal Heartbeat. To improve interpretability, Grad-CAM is applied to the best-performing model to visualize the ECG regions contributing most to the final decision. Additional analyses are also conducted to examine explanation stability, regional sensitivity, the contribution of transfer learning, and confidence-based reliability.

\section{Related works}

Several studies have tackled automated ECG interpretation by converting signals into images and applying DL classifiers such as CNN, RNN, LSTM etc. \cite{ 11493872}. Ruan et al. \cite{ruan2022arrhythmia} proposed transforming ECG signals into time-frequency spectrograms for a 2D CNN improved arrhythmia detection accuracy and showed that image-based ECG classification can work well with convolutional models. However, the method remained a “black box” and focused mainly on single-lead data. Later studies explored multi-lead image representations to capture inter-lead features. For example, Li et al. \cite{li2022two} converted 12-lead recordings into a 2D grayscale image and used a specialized ResNet to learn intra- and inter-lead patterns, improving classification of 8 arrhythmia types. Although these image-based methods achieved high accuracy, they initially lacked explainability, which limited clinical use.

To improve interpretability, recent studies have integrated explainable AI into ECG image classification, often using visual attention maps like Grad-CAM to highlight image regions that drive the model’s prediction. Martono et al. \cite{10.1145/3589437.3589443} applied a CNN to ECG recurrence plot images and used Grad-CAM to highlight salient regions for different arrhythmias, enabling comparison between the model’s focus and clinicians’ diagnostic cues. Large-scale studies have shown that explainability improves model credibility. Trained a CNN on over two million annotated ECG charts for multi-label classification and used Grad-CAM to verify that its predictions were based on medically relevant waveform features, such as lead segments linked to arrhythmias \cite{sangha2022automated}. Likewise, Murat et al. \cite{murat2026mm} introduced a multimodal approach (MM-GradCAM) that processes both 1D signal and 2D image inputs, producing separate Grad-CAM explainability maps for each. 

Beyond CAM-based post-hoc explainability, researchers have also used model-agnostic methods on ECG classifiers. One study applied LIME to identify which ECG segments most influenced a CNN’s heartbeat classification, providing case-specific explanations for its predictions \cite{gliner2025clinically}. This approach produced heatmaps that closely matched physicians’ annotations of pathological patterns. Despite this progress, challenges remain. Overall, explainable image-based ECG classification methods, from attention maps to local surrogate models, have advanced, but each still faces limits such as partial clinical alignment or implementation complexity. %These gaps motivate the present work, which focuses on a more generalizable and interpretable framework for ECG image classification, leveraging insights from prior methods (Grad-CAM, attention mechanisms, and model-agnostic tools) to improve both transparency and diagnostic reliability.

\section{Methods and Materials}

\begin{figure}[!t]
\centering
\includegraphics[width=.8\linewidth]{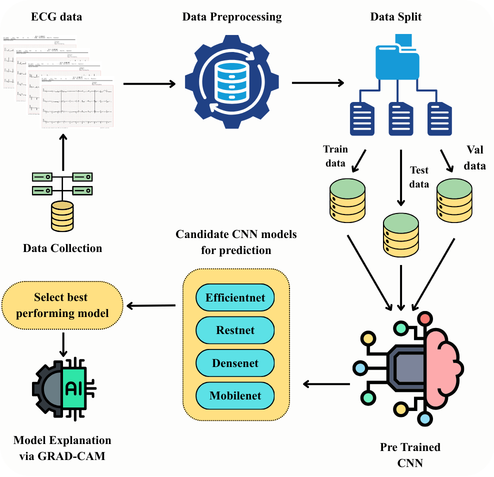}
\caption{Block diagram of the proposed transfer-learning and explainability framework for ECG image classification. ECG images are collected, preprocessed, and split into training, validation, and test sets. Multiple pretrained CNN backbones are then fine-tuned and compared under a common pipeline, the best model is selected, and Grad-CAM is applied to highlight the image regions most responsible for the final class decision.}
\label{fig:method}
\end{figure}

The proposed methodology has been summerized in Fig.~\ref{fig:method}. The pipeline begins with data collection and proceeds through data preprocessing, model training and lastly model explanation.

\FloatBarrier

\subsection{Dataset and Preprocessing}

A publicly available (clinically validated) ECG image dataset was used in this study which was collected from Ch. Pervaiz Elahi Institute of Cardiology, Multan, Pakistan \cite{spiritos2024ecg}. The dataset contains 928 images of ecg data with four classes:

\begin{itemize}
    \item \textbf{Normal Heartbeat:} 284 image
    \item \textbf{Myocardial Infarction:} 239 image
    \item \textbf{Abnormal Heartbeat:} 233 iamge
    \item \textbf{History of Myocardial Infarction:} 172 image
\end{itemize}

Preprocessing consisted of removing overlaid text artifacts, applying a custom crop to suppress non-signal margins, converting each image to three-channel grayscale format, resizing to $224 \times 224$ pixels, and normalizing using ImageNet mean and standard deviation values. During training, data augmentation included brightness and contrast jitter, together with mild affine transformations consisting of rotation ($\pm 3^\circ$), translation (up to 5\%), and scaling (0.95-1.05), in order to improve invariance and reduce overfitting. The dataset was first divided into training-validation and test subsets using stratified sampling, after which the training-validation subset was further split into training and validation partitions. The final split ratio was approximately 70\% training, 10\% validation, and 20\% testing.
 
\subsection{Fine-Tuning Protocol for CNN Pretrained Models}

A unified transfer learning protocol was used to compare multiple ImageNet-pretrained CNN backbones, including EfficientNet-B0, ResNet-50, DenseNet-121, and MobileNetV3-Large. For each model, the original task-specific classifier was replaced with a new \texttt{NUM\_CLASSES} output layer, with dropout added where appropriate. Training followed a two-stage strategy, first, the feature extractor was frozen and only the classification head was trained using AdamW, with model selection based on the best validation macro-F1, second, the last 35\% of parameters were unfrozen for partial fine-tuning with a lower learning rate and early stopping (patience = 6) to reduce overfitting while adapting higher-level features. Cross-entropy loss was used, with an optional class-weighted variant to address class imbalance, and mixed precision training (autocast with gradient scaling) was applied for GPU efficiency. 

To examine the effect of transfer learning, two additional baselines were evaluated using the same data split and preprocessing pipeline, a shallow custom CNN trained from scratch, and a ResNet-50 initialized without ImageNet-pretrained weights. Both baselines were trained with the same loss function and assessed using the same test metrics.

\subsection{Evaluation Metrics}
Model performance was assessed using standard classification metrics, including accuracy, precision, recall, F1-score, and the confusion matrix. Accuracy quantifies the overall proportion of correctly classified samples. Precision reflects the reliability of positive predictions by measuring the fraction of predicted positives that are true positives, whereas recall captures sensitivity by measuring the fraction of true positives correctly identified. The F1-score summarizes the balance between precision and recall via their harmonic mean. In addition, the confusion matrix provides a class-wise breakdown of predictions, reporting counts for each true class against each predicted class. Lastly, The ROC curve is used that plots the true positive rate against the false positive rate across decision thresholds.

\begin{equation}
\mathrm{Accuracy}=\frac{\text{Number of Correct Predictions}}{\text{Total Number of Predictions}}
\label{eq:acc}
\end{equation}

\begin{equation}
\mathrm{Precision}=\frac{TP}{TP+FP}
\label{eq:prec}
\end{equation}

\begin{equation}
\mathrm{Recall}=\frac{TP}{TP+FN}
\label{eq:rec}
\end{equation}

\begin{equation}
F_{1}=\frac{2 * \,\mathrm{Precision}\,\mathrm{*  Recall}}{\mathrm{Precision}+\mathrm{Recall}}
\label{eq:f1}
\end{equation}

%In addition to standard classification metrics, supplementary analyses were performed to assess explanation behavior and prediction reliability. Grad-CAM stability under randomization was quantified using correlation and IoU@20\% between saliency maps from the trained model and a randomly initialized model of the same architecture. Regional sensitivity was evaluated by measuring the drop in prediction confidence after masking either border regions or the central image region. Reliability was further examined using the mean confidence of correct and incorrect predictions, together with the count of high-confidence misclassifications.

\subsection{Model Explainability}

Model interpretability was examined using a PyTorch implementation of Grad-CAM applied to the final convolutional layer of the selected CNN backbone. Forward activations and backward gradients for the target class were captured through hooks, and channel-wise importance weights were obtained by globally averaging the gradients. A class-discriminative heatmap was then generated from the weighted feature maps, passed through ReLU, normalized to [0,1], resized to the input image size, and overlaid on the ECG image. For qualitative analysis, representative correct and incorrect test predictions were visualized with their corresponding Grad-CAM maps to inspect whether the model focused on waveform-relevant regions and to identify potential failure cases.

\subsection{Additional Empirical Analyses}

To complement the classification results and Grad-CAM visualizations, we performed four additional analyses. First, we assessed explanation stability by comparing Grad-CAM maps from the trained ResNet-50 with maps from a randomly initialized ResNet-50 using correlation and IoU@20\% agreement. Second, we conducted regional sensitivity analysis by masking border regions and the central waveform-rich region and measuring the resulting confidence drop. Third, we evaluated the role of transfer learning using two baselines, a shallow custom CNN and a randomly initialized ResNet-50. Finally, we analyzed confidence-based reliability by comparing confidence scores for correct and incorrect predictions and counting high-confidence errors on the test set.

\section{Result Analysis}

All experiments were carried out in Jupyter Notebook on a Windows 11 laptop equipped with an Intel Core i5 processor, 16 GB RAM and a 1650 Nvidia GTX GPU. For experiemtnal analysis python programming language was used and libraries like pytorch, cv2, torchvision etc were utilized. 

\subsection{Performance evalatuation metrics}

Across the evaluated CNN backbones, ResNet-50 achieved the strongest overall performance, reaching an accuracy of 0.9194 with a macro-F1 of 0.9098 (Table~\ref{tab:cnn_backbone_classwise}). DenseNet-121 followed closely (accuracy 0.9032, macro-F1 0.8925), indicating competitive generalization across classes. In contrast, EfficientNet-B0 achieved moderate performance (accuracy 0.8172, macro-F1 0.8174), while MobileNetV3-Large got the lowest overall results (accuracy 0.7204, macro-F1 0.7114), suggesting a larger drop in balanced class performance.

Class-wise behavior shows clear differences in error patterns. Myocardial Infarction was detected with very high sensitivity by ResNet-50, DenseNet-121, and MobileNetV3-Large (recall = 1.0000), with ResNet-50 yielding the highest F1 for this class (0.9600). Normal Heartbeat was recovered almost perfectly by ResNet-50 and DenseNet-121 (recall = 1.0000, F1 = 0.9344 and 0.9580, respectively), whereas MobileNetV3-Large struggled substantially (recall = 0.4912, F1 = 0.6154), which largely explains its lower macro scores. History of Myocardial Infarction remained comparatively challenging across models, with the best F1 observed for DenseNet-121 (0.8406) and reduced robustness for MobileNetV3-Large (0.6479). Finally, Abnormal Heartbeat exhibited consistently high precision (often near 1.0000) but variable recall, indicating that some backbones were conservative in assigning this label, trading missed detections for fewer false positives.

\begin{table*}[!t]
\centering
\caption{Class-wise precision, recall, and F1-score of different CNN backbones, with macro-averaged metrics and overall accuracy.}
\label{tab:cnn_backbone_classwise}
\scriptsize
\renewcommand{\arraystretch}{1.25}
\setlength{\tabcolsep}{5pt}
\begin{tabular}{l l c c c c c c c}
\hline
\textbf{Model} & \textbf{Class} & \textbf{Precision} & \textbf{Recall} & \textbf{F1-score} & \textbf{Macro Precision} & \textbf{Macro Recall} & \textbf{Macro F1 Score} & \textbf{Accuracy} \\
\hline

\multirow{4}{*}{ResNet-50}
& Abnormal Heartbeat                 & 1.0000 & 0.8511 & 0.9195 & \multirow{4}{*}{0.9241} & \multirow{4}{*}{0.9039} & \multirow{4}{*}{0.9098} & \multirow{4}{*}{0.9194} \\
& History of Myocardial Infarction   & 0.8966 & 0.7647 & 0.8254 \\
& Myocardial Infarction              & 0.9231 & 1.0000 & 0.9600 \\
& Normal Heartbeat                   & 0.8769 & 1.0000 & 0.9344 \\
\cline{1-9}

\multirow{4}{*}{DenseNet-121}
& Abnormal Heartbeat                 & 1.0000 & 0.7234 & 0.8395 & \multirow{4}{*}{0.9052} & \multirow{4}{*}{0.8941} & \multirow{4}{*}{0.8925} & \multirow{4}{*}{0.9032} \\
& History of Myocardial Infarction   & 0.8286 & 0.8529 & 0.8406 \\
& Myocardial Infarction              & 0.8727 & 1.0000 & 0.9320 \\
& Normal Heartbeat                   & 0.9194 & 1.0000 & 0.9580 \\
\cline{1-9}

\multirow{4}{*}{EfficientNet-B0}
& Abnormal Heartbeat                 & 0.8000 & 0.8511 & 0.8247 & \multirow{4}{*}{0.8194} & \multirow{4}{*}{0.8162} & \multirow{4}{*}{0.8174} & \multirow{4}{*}{0.8172} \\
& History of Myocardial Infarction   & 0.8438 & 0.7941 & 0.8182 \\
& Myocardial Infarction              & 0.8125 & 0.8125 & 0.8125 \\
& Normal Heartbeat                   & 0.8214 & 0.8070 & 0.8142 \\
\cline{1-9}

\multirow{4}{*}{MobileNetV3-Large}
& Abnormal Heartbeat                 & 0.8537 & 0.7447 & 0.7955 & \multirow{4}{*}{0.7369} & \multirow{4}{*}{0.7281} & \multirow{4}{*}{0.7114} & \multirow{4}{*}{0.7204} \\
& History of Myocardial Infarction   & 0.6216 & 0.6765 & 0.6479 \\
& Myocardial Infarction              & 0.6486 & 1.0000 & 0.7869 \\
& Normal Heartbeat                   & 0.8235 & 0.4912 & 0.6154 \\
\hline

\end{tabular}
\end{table*}

\subsection{Quantitative Results and Error Analysis}

%Fig.~\ref{fig:lc} and show the training behavior across epochs. The loss decreases steadily for the training, validation, and test splits, and the three curves remain close for most of training, indicating stable convergence with a limited generalization gap. After the mid training phase, the improvements become smaller and the curves begin to flatten, suggesting that the model is nearing its performance limit on this dataset.

\begin{figure}
\label{acc_curve}
    \centering
    \includegraphics[width=.9\linewidth]{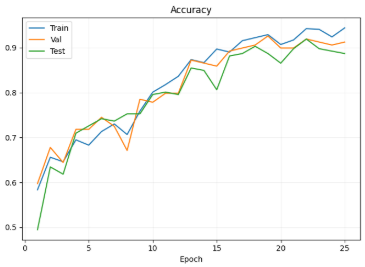}
    \caption{Training, validation, and test accuracy curves over 25 epochs for the selected ECG classification model.}
    \label{fig:ac}
\end{figure}

Fig.~\ref{fig:ac} represents training, validation, and test accuracy curves over 25 epochs for the selected ECG classification model. The accuracy curves show a similar trend, increasing consistently before plateauing near the final epochs, while validation and test accuracy closely follow the training curve. Overall, this pattern suggests effective learning without clear overfitting, since validation and test performance improve alongside training rather than dropping as training accuracy continues to increase.

%\begin{figure}
%\label{loss_curve}
 %   \centering
  %  \includegraphics[width=.9\linewidth]{train test loss.png}
   % \caption{Training, validation, and test loss curves across 25 epochs for the selected ECG classification model.}
   % \label{fig:lc}
%\end{figure}

\begin{figure}
\label{roc}
    \centering
    \includegraphics[width=.9\linewidth]{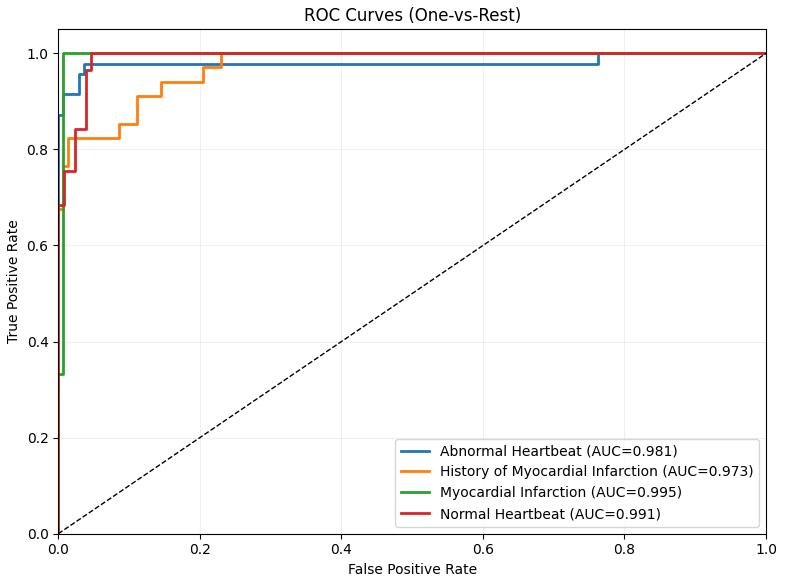}
    \caption{One-vs-rest ROC curves for the four ECG classes. The classifier shows strong discriminative performance across all classes.}
    \label{fig:rc}
\end{figure}

\begin{figure}
\label{fig:acc_curve}
    \centering
    \includegraphics[width=1\linewidth]{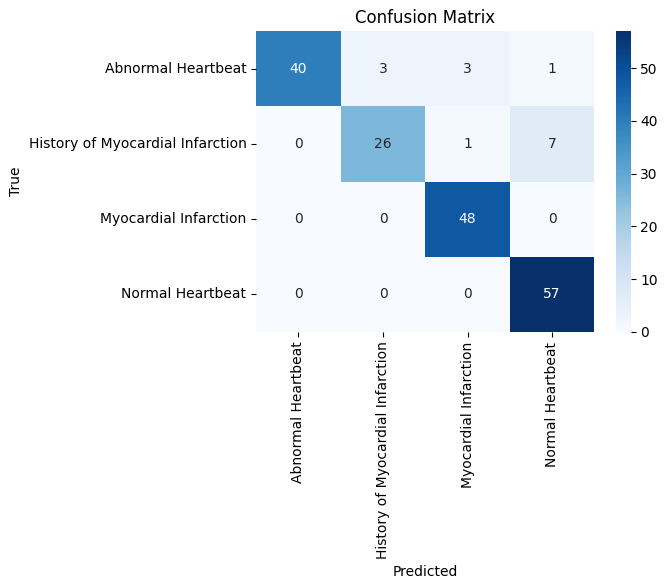}
    \caption{Confusion matrix of the four-class ECG classifier on the test set.}
    \label{fig:cf}
\end{figure}

Fig.~\ref{fig:rc} shows one-vs-rest ROC curves with consistently high separability for all classes, Myocardial Infarction (AUC = 0.995) and Normal Heartbeat (AUC = 0.991) exhibit near-ideal discrimination, while Abnormal Heartbeat (AUC = 0.981) and History of Myocardial Infarction (AUC = 0.973) remain strongly separable but relatively more challenging. The confusion matrix in Fig.~\ref{fig:cf} explains this pattern, Myocardial Infarction (48/48) and Normal Heartbeat (57/57) are classified perfectly, whereas most errors concentrate in History of Myocardial Infarction, which is primarily confused with Normal Heartbeat (7 samples) and occasionally with Myocardial Infarction (1 sample). For Abnormal Heartbeat, misclassifications are distributed across History MI (3), MI (3), and Normal (1), indicating partial feature overlap with adjacent conditions. Overall, the matrix yields 171 correct predictions out of 186 (accuracy 0.9194), aligning with the reported aggregate metrics and confirming that performance limitations are driven mainly by the History of MI vs Normal boundary rather than global model instability.

\subsection{Grad-CAM on predictions}

\begin{figure}[]
    \centering
    \includegraphics[width=.75\linewidth]{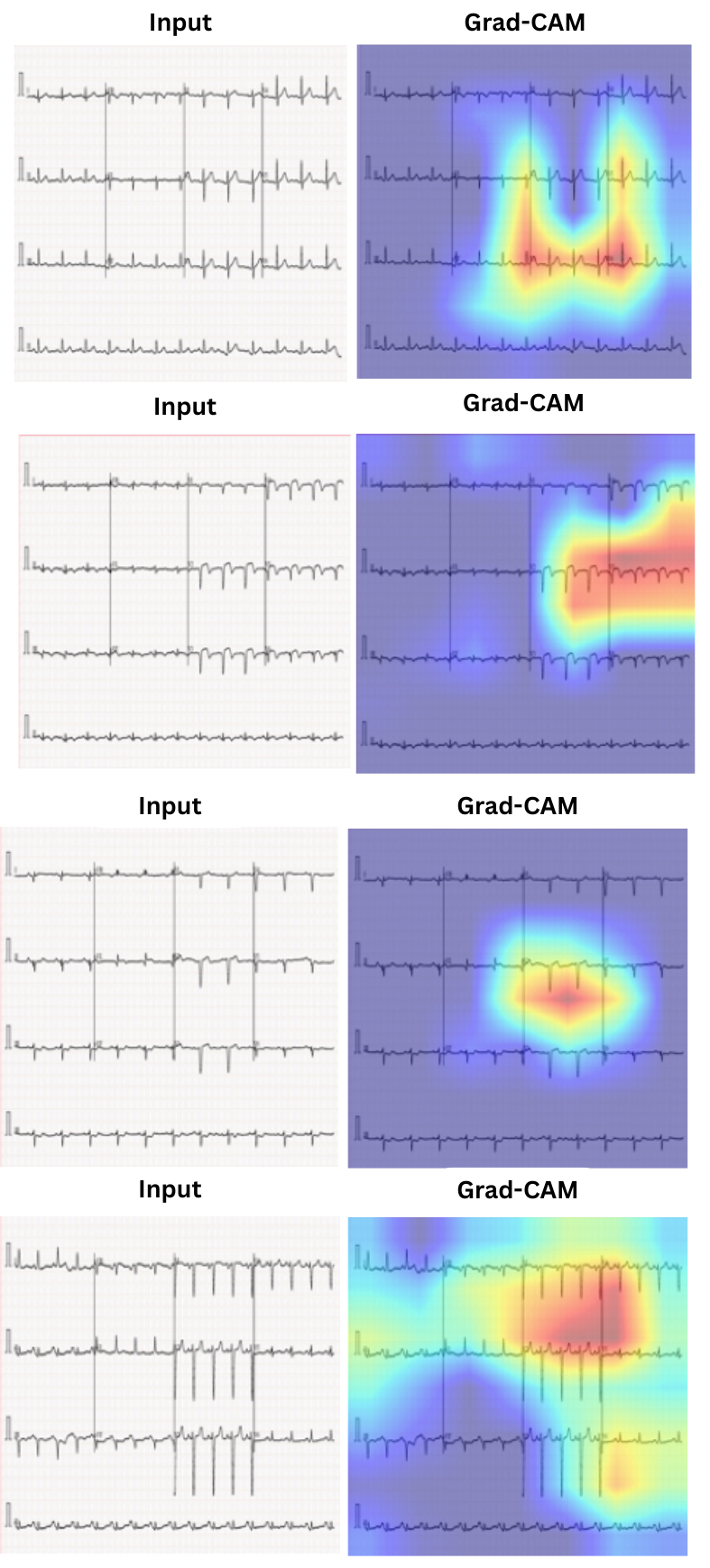}
    \caption{Representative input ECG images and corresponding Grad-CAM overlays for four test examples from different diagnostic classes. For each case, the heatmap localizes the waveform regions and lead segments that contribute most to the model’s prediction (warmer colors indicate higher contribution). The examples illustrate class-specific attention patterns, with concentrated activation over discriminative complexes and segment-level abnormalities that drive the final classification.}
    \label{fig:gradcam}
\end{figure}

These Grad-CAM visualizations provide a model-driven explanation of which ECG regions most influenced each correct prediction (warm colors indicate higher contribution, cooler colors indicate lower contribution). Across the examples, the highlighted areas align primarily with waveform morphology rather than the background grid or margins, indicating that the classifier is leveraging signal-relevant structures. For the Normal Heartbeat case, the activation concentrates around repeated, well-formed complexes, consistent with the model verifying stable rhythmic patterns and clean morphology instead of reacting to noise-like regions. This supports that the decision is grounded in the ECG trace itself, not spurious layout cues. Fig.~\ref{fig:gradcam} represents a sample of 4 input images and 4 Grad-CAM overlays from the test set.

%\begin{figure*}[!t]
%  \centering
%  \includegraphics[width=30.pc]{input grad cam.png}
%  \caption{}
%  \label{fig:workflow}
%\end{figure*}

For History of Myocardial Infarction, the saliency becomes more localized to specific segments of the trace, emphasizing regions where subtle morphology differences are expected to appear (changes in QRS shape and surrounding repolarization patterns). The two History-of-MI examples show slightly different but still focused attention, suggesting the model adapts to case-specific discriminative cues rather than using a single fixed template. In the Abnormal Heartbeat example, the activation broadens and intensifies over segments containing irregular or atypical waveform behavior, consistent with the model prioritizing areas that likely capture abnormal rhythm/morphology characteristics. Overall, these Grad-CAM maps strengthen interpretability by demonstrating that predictions are driven by clinically meaningful waveform regions, and they also provide a practical tool for auditing failure modes by checking whether attention shifts to irrelevant areas when misclassifications occur.

\subsubsection{Additional Empirical Evaluation}

To strengthen the empirical assessment beyond standard classification metrics, we performed four additional analyses such as an explanation stability check based on Grad-CAM randomization, a regional sensitivity analysis, stronger baseline comparisons, and a confidence-based reliability analysis.

\paragraph{Explanation Stability Check}
We first examined whether the Grad-CAM maps reflected learned representations rather than architecture-induced artifacts. For this purpose, we compared saliency maps produced by the trained ResNet-50 with those generated by a randomly initialized model of the same architecture. The agreement between trained and random Grad-CAM maps was low, with a mean correlation of $-0.038$ and a mean IoU@20\% of $0.139$. This result suggests that the saliency patterns produced by the trained model were shaped by learned parameters rather than being solely driven by the architectural structure of the network.

\paragraph{Regional Sensitivity Analysis}
We further evaluated whether the classifier relied primarily on waveform-rich ECG regions or on broader peripheral structure. To do this, we separately masked border regions and the central image region, and measured the reduction in prediction confidence. Removing the central region caused a slightly larger mean confidence drop than removing the border regions ($0.599$ vs.\ $0.559$), indicating that the model relied somewhat more on central waveform-dense content than on peripheral context. However, the relatively small gap suggests that broader image structure may still contribute to the prediction process.

\paragraph{Transfer Learning versus Training from Scratch}
To better quantify the value of transfer learning, we introduced two additional baselines, a shallow custom CNN trained from scratch and a ResNet-50 trained from scratch without ImageNet initialization. Both scratch models collapsed to near-majority-class behavior, each achieving only $0.253$ test accuracy and $0.101$ macro-F1. In contrast, the pretrained ResNet-50 achieved $0.919$ test accuracy and $0.911$ macro-F1. This large performance gap demonstrates that pretrained representations are critical for ECG image classification under limited-data conditions.

\paragraph{Confidence-Based Reliability Analysis}
Finally, we analyzed predictive confidence to better characterize reliability. On the test set, correct predictions had a substantially higher mean confidence than incorrect predictions ($0.768$ vs.\ $0.597$). Among the $27$ incorrect predictions, only one exceeded a confidence threshold of $0.90$. This indicates that most model errors occurred under comparatively lower certainty, although isolated high-confidence mistakes remained present.

\begin{table}[t]
\caption{Summary of the additional empirical analyses. The table reports Grad-CAM explanation stability under randomization, confidence changes under regional masking, performance differences between pretrained and from-scratch baselines, and confidence-based reliability statistics on the test set. These results complement the main classification metrics by providing further evidence on explanation behavior, spatial sensitivity, transfer learning effects, and predictive reliability.}
\label{tab:additional_analysis_short}
\centering
\scriptsize
%\footnotesize
\setlength{\tabcolsep}{3pt}
\renewcommand{\arraystretch}{1.05}
\begin{tabular}{lll}
\hline
\textbf{Category} & \textbf{Metric} & \textbf{Value} \\
\hline
Explanation stability & Corr. / IoU@20\% & $-0.038$ / $0.139$ \\
Regional sensitivity & Border / center drop & $0.559$ / $0.599$ \\
Transfer learning & Pretrained ResNet-50 acc. / F1 & $0.919$ / $0.911$ \\
Transfer learning & SimpleECGCNN acc. / F1 & $0.253$ / $0.101$ \\
Transfer learning & ResNet-50 scratch acc. / F1 & $0.253$ / $0.101$ \\
Confidence reliability & Mean conf. corr. / incorr. & $0.768$ / $0.597$ \\
Confidence reliability & Incorrect preds. / high-conf. errors & $27$ / $1$ \\
\hline
\end{tabular}
\end{table}

\section{Conclusion}

This study presented an explainability-aware transfer learning framework for four-class ECG image classification, including Abnormal Heartbeat, History of Myocardial Infarction, Myocardial Infarction, and Normal Heartbeat. Among the evaluated pretrained CNN backbones, ResNet-50 achieved the best overall performance, reaching 91.94\% accuracy and 0.9098 macro-F1, with strong ROC separability across all classes. Error analysis showed that the main challenge lies in distinguishing History of Myocardial Infarction from Normal Heartbeat, while Myocardial Infarction and Normal Heartbeat were classified reliably on the reported test set. Grad-CAM visualizations and supplementary analyses provided supportive evidence that the model relied substantially on waveform-relevant regions, although some contribution from broader image context may still remain. In addition, the large gap between pretrained and from-scratch baselines highlighted the importance of transfer learning under limited-data conditions. Despite these promising results, the study remains limited by dataset scale, class difficulty, and the lack of external multi-source validation. 

This work does not aim to replace signal-based ECG interpretation. Instead, it focuses on ECG image classification for settings where only printed, scanned, or image-format ECG records are available, such as retrospective record analysis, low-resource clinical workflows, and explainability-aware decision-support benchmarking. Future work should evaluate the framework on larger and more diverse ECG image datasets, include stronger quantitative assessment of explanation quality, and further improve robustness through techniques such as class-imbalance handling, domain adaptation, and confidence-aware prediction. These steps would provide a stronger basis for assessing real-world generalization and practical clinical usefulness. From a systems perspective, this work contributes toward the development of trustworthy AI solutions for healthcare by combining predictive performance with explanation-based validation. These capabilities support model auditing, improve transparency, and enable safer integration into clinical decision support systems.

\bibliographystyle{IEEEtranS}

\bibliography{biblio_References}

@article{BALOGLU201923,
title = {Classification of myocardial infarction with multi-lead ECG signals and deep CNN},
journal = {Pattern Recognition Letters},
volume = {122},
pages = {23-30},
year = {2019},
issn = {0167-8655},
doi = {https://doi.org/10.1016/j.patrec.2019.02.016},
url = {https://www.sciencedirect.com/science/article/pii/S016786551930056X},
author = {Ulas Baran Baloglu and Muhammed Talo and Ozal Yildirim and Ru San Tan and U Rajendra Acharya}
}

@ARTICLE{11493872,
  author={Tahsin, Mohammad Sadman and Adarbah, Haitham Y. and Noore, Afzel},
  journal={IEEE Access}, 
  title={A Lightweight Channel-Attention CNN for Robust Beat-Level Arrhythmia Detection}, 
  year={2026},
  volume={14},
  number={},
  pages={62803-62817},
  doi={10.1109/ACCESS.2026.3686810}}

@article{sangha2022automated,
  title={Automated multilabel diagnosis on electrocardiographic images and signals},
  author={Sangha, Veer and Mortazavi, Bobak J and Haimovich, Adrian D and Ribeiro, Ant{\^o}nio H and Brandt, Cynthia A and Jacoby, Daniel L and Schulz, Wade L and Krumholz, Harlan M and Ribeiro, Antonio Luiz P and Khera, Rohan},
  journal={Nature communications},
  volume={13},
  number={1},
  pages={1583},
  year={2022},
  doi = {doi.org/10.1038/s41467-022-29153-3},
  publisher={Nature Publishing Group UK London}
}

@Article{e23010018,
AUTHOR = {Linardatos, Pantelis and Papastefanopoulos, Vasilis and Kotsiantis, Sotiris},
TITLE = {Explainable AI: A Review of Machine Learning Interpretability Methods},
JOURNAL = {Entropy},
VOLUME = {23},
YEAR = {2021},
NUMBER = {1},
ARTICLE-NUMBER = {18},
URL = {https://www.mdpi.com/1099-4300/23/1/18},
PubMedID = {33375658},
ISSN = {1099-4300},
DOI = {10.3390/e23010018}
}

@article{goettling2024xecgarch,
  title={xECGArch: a trustworthy deep learning architecture for interpretable ECG analysis considering short-term and long-term features},
  author={Goettling, Marc and Hammer, Alexander and Malberg, Hagen and Schmidt, Martin},
  journal={Scientific Reports},
  volume={14},
  number={1},
  pages={13122},
  year={2024},
  publisher={Nature Publishing Group UK London}
}

@article{ruan2022arrhythmia,
  title={Arrhythmia classification and diagnosis based on ECG signal: A multi-domain collaborative analysis and decision approach},
  author={Ruan, Hongpeng and Dai, Xueying and Chen, Shengqi and Qiu, Xiang},
  journal={Electronics},
  volume={11},
  number={19},
  pages={3251},
  year={2022},
  publisher={MDPI}
}

@article{li2022two,
  title={Two-dimensional ECG-based cardiac arrhythmia classification using DSE-ResNet},
  author={Li, Jiahao and Pang, Shao-peng and Xu, Fangzhou and Ji, Peng and Zhou, Shuwang and Shu, Minglei},
  journal={Scientific Reports},
  volume={12},
  number={1},
  pages={14485},
  year={2022},
  publisher={Nature Publishing Group UK London}
}

@inproceedings{10.1145/3589437.3589443,
author = {Martono, Niken Prasasti and Nishiguchi, Toru and Ohwada, Hayato},
title = {Interpreting Arrhythmia Classification Using Deep Neural Network and CAM-Based Approach},
year = {2023},
isbn = {9781450397636},
publisher = {Association for Computing Machinery},
address = {New York, NY, USA},
url = {https://doi.org/10.1145/3589437.3589443},
doi = {10.1145/3589437.3589443},
booktitle = {Proceedings of the 2022 6th International Conference on Computational Biology and Bioinformatics},
pages = {35–40},
numpages = {6},
location = {Bali Island, Indonesia},
series = {ICCBB '22}
}

@article{murat2026mm,
  title={MM-GradCAM: an improved multimodal GradCAM method with 1D and 2D ECG data for detection of cardiac arrhythmia},
  author={Murat Duranay, Fatma and Murat, Ender and Y{\i}ld{\i}r{\i}m, {\"O}zal and Demir, Yakup and Tan, Ru-San and Sampathila, Niranjana and Acharya, U Rajendra},
  journal={Scientific Reports},
  year={2026},
  publisher={Nature Publishing Group UK London}
}

@article{gliner2025clinically,
  title={Clinically meaningful interpretability of an AI model for ECG classification},
  author={Gliner, Vadim and Levy, Idan and Tsutsui, Kenta and Acha, Moshe Rav and Schliamser, Jorge and Schuster, Assaf and Yaniv, Yael},
  journal={NPJ Digital Medicine},
  volume={8},
  number={1},
  pages={109},
  year={2025},
  publisher={Nature Publishing Group UK London}
}

@misc{spiritos2024ecg,
  author       = {E. Spiritos},
  title        = {ECG Images Dataset of Cardiac Patients},
  year         = {2024},
  howpublished = {Kaggle dataset},
  note         = {Available online},
  url          = {https://www.kaggle.com/datasets/evilspirit05/ecg-analysis}
}
%\section*{Acknowledgment}

\vspace{12pt}
%\color{red}
%IEEE conference templates contain guidance text for composing and formatting conference papers. Please ensure that all template text is removed from your conference paper prior to submission to the conference. Failure to remove the template text from your paper may result in your paper not being published.

\end{document}